# Deep Thermal Imaging: Proximate Material Type Recognition in the Wild through Deep Learning of Spatial Surface Temperature Patterns

**Youngjun Cho**    **Nadia Bianchi-Berthouze**    **Nicolai Marquardt**    **Simon J. Julier**
University College London, London, United Kingdom
{youngjun.cho.15, nadia.berthouze, n.marquardt, s.julier}@ucl.ac.uk

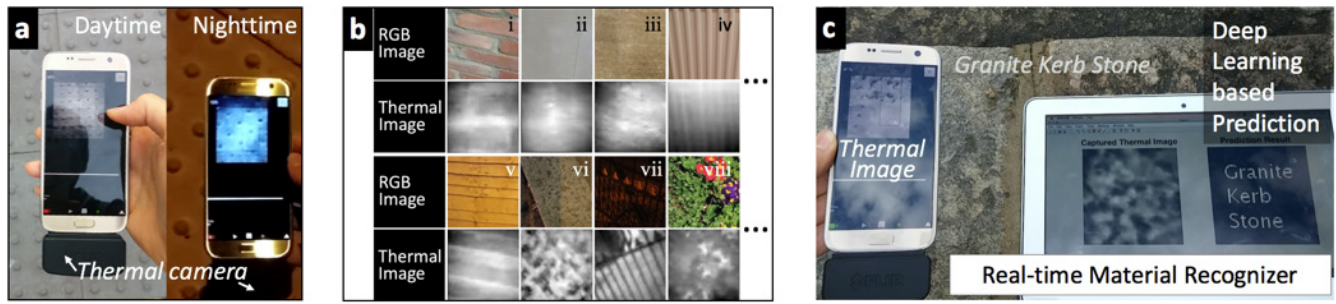

**Figure 1. Deep Thermal Imaging for in-the-wild material recognition: (a) spatial temperature patterns of material during day/nighttime, (b) examples of RGB and quantized thermal images of materials (i: brick; ii: marble; iii: carpet; iv: radiator; v: wood fence; vi: granite; vii: iron fence; viii: flower bed), (c) the real-time material recognizer.**

## ABSTRACT
We introduce *Deep Thermal Imaging*, a new approach for close-range automatic recognition of materials to enhance the understanding of people and ubiquitous technologies of their proximal environment. Our approach uses a low-cost mobile thermal camera integrated into a smartphone to capture thermal textures. A deep neural network classifies these textures into material types. This approach works effectively without the need for ambient light sources or direct contact with materials. Furthermore, the use of a deep learning network removes the need to handcraft the set of features for different materials. We evaluated the performance of the system by training it to recognize 32 material types in both indoor and outdoor environments. Our approach produced recognition accuracies above 98% in 14,860 images of 15 indoor materials and above 89% in 26,584 images of 17 outdoor materials. We conclude by discussing its potentials for real-time use in HCI applications and future directions.

## ACM Classification Keywords
H.5.2. Information interfaces and presentation (e.g., HCI): User Interfaces - Input devices and strategies

## INTRODUCTION
Context awareness, including material type awareness, is of critical importance in many situations for both humans and ubiquitous technologies [13,15,49]. Methods that contribute to build context awareness can support decision making, increase safety and facilitate interaction with the environment. A compelling use case is Microsoft's *Seeing AI* [58] for visually impaired people. Using a smartphone camera and microphone, *Seeing AI* recognizes the environment and provides an audio description of the immediate context (e.g., the scene and people nearby). While such systems are useful, they do not provide all information the users may need. For example, in the case of physical manipulation, it is not sufficient to know what the object is, but also what material it is made of.

Creating the ability to recognize materials is the focus of our paper. A variety of sensing technology have been investigated in HCI to support material awareness. SpecTrans [48], for instance, is an image sensor-based material recognizer that analyzes reflected light patterns produced by multi-spectra lighting sources. Among its use cases, the authors show how it could support embedded marker applications (e.g., encoded reality [59]) without the need to physically modify the objects. Another example is RadarCat [55], a contact-based recognizer which uses a lightweight radar system to develop a "material dictionary", allowing users to search for the name of the material an object is made of by simply touching it with the recognizer. Whilst these applications enable new ways to understand and interact with materials, they require controlled illumination and direct contact with those materials, which is not always possible or desirable in real life situations.

To overcome the above limitations, we propose a new proximate (10cm-50cm range) material recognition system which uses a smartphone-based mobile thermal camera and a deep learning technique to classify material types. This approach, illustrated in Figure 1, requires no physical contact between the sensor and materials and is robust to environmental light and temperature changes. In addition, by using deep learning, we remove the need to hand-engineer features for recognition. Such techniques make it easier to extend the recognition capabilities to new types of materials by simply adding sufficient new training data [18,57]. This is critical for real-world applications where the materials to be recognized are not always all known in advance.

**Contributions -** This paper makes four main contributions:

- *Deep Thermal Imaging*: a novel material recognizer system based on a low-cost, mobile thermal camera which can be integrated into a smartphone and is robust to changes in illumination and temperature conditions.

- Deep learning-based software tools for feature learning on spatial thermal textures of materials, enabling easy extension to a wider set of materials. The implementation of our recognition engine is open-sourced.

- In-the-wild studies with 32 types of indoor and outdoor materials (for a total of 41,444 thermal images) to demonstrate the capability and robustness of Deep Thermal Imaging to discriminate between materials under different (i) illumination settings, (ii) environment temperatures, (iii) point of views and proximate distances.

- Discussion of practical HCI use cases benefitting from *Deep Thermal Imaging* material recognition capabilities.

**RELATED WORK**
In this section, we discuss prior work on *proximity sensing and its use in context-aware HCI applications*, review technical methods for *material classification* and synthesize studies about *thermal imaging*.

**Proximity Sensing in HCI Applications**
Proximity is defined as "physical distance" in social contexts [12]. The term *proximity sensing* is generally use to indicate the ability to detect the presence of close-range objects [6]. Various types of sensors have been used for proximity sensing, including ultrasonic sensors [33], capacitive sensors [54], laser range finders [42] and photo-detectors [21]. Proximity sensing is useful in many HCI situations. For example, Eric and Jochen [16] explored how detecting the presence of nearby objects can help construction workers avoid potential hazards. Boyd et al. [3] proposed a necklace-type wearable prototype using photo-detector to help people with autism gain awareness of their surroundings. Computer agents have also exploited proximity sensing to interact with people, e.g., to understand people's in-air gesture and hovering over a mouse [22] or a display screen [9,60]. Whilst this technology has been used to increase context awareness, its focus has been mainly on a) simply signaling that an object is present or b) detecting the presence of well pre-defined object (often human skin). Wiese et al. [54] have started to enhance proximity sensing capability by combining multiple sensing modalities to sense the type of material a mobile phone is placed on. However, their range of materials is limited to carry out predefined tasks such as whether the phone is in a pocket.

**Material Classification using Sensing Devices**
The ability to automatically recognize materials has been recently explored by HCI and computer vision researchers. The aims have been to enable novel ubiquitous interaction and richer context awareness [13,15,49]. For instance, Harrison and Hudson [20] have prototyped a photoresistor based material recognizer. Sato et al. [48] have proposed a real-time transparent and specular material type recognizer using a mouse image sensor and light emitters. Yeo et al. [55] have implemented a low-cost radar system-based material recognizer, and have used a smartphone front-camera and screen flash light [56] for better scalability. Researchers have also explored how friction and acceleration signals can be used to understand or recognize materials [7,31,35].

Whilst these approaches support material awareness, they have some limitations. The most important is that many methods require the sensors to be in contact with the material or very close to it (with 10cm of). In many situations, this is not feasible (e.g., hazardous, or is not socially acceptable).

Visible spectrum-based imaging approaches (conventional RGB) [2,14,34,53], on the other hand, can provide touchless ways to discriminate between different materials. However, they are susceptible to *illumination effects*: for instance, extremes of darkness and brightness incapacitate these algorithms due to sensor saturation or sensitivity [1,34].

This limitation can be addressed through the use of other imaging modalities that read electromagnetic radiation outside the visible range (e.g., radar imaging [5], thermal imaging [36,52]). In fact, the human eyes and RGB cameras that mimic the eyes are only sensitive to a narrow range of electromagnetic spectrum called "visible light". They capture objects through the radiations in the visible spectrum their surface emits or reflects when hit by a source of light [45]. This means that without sources emitting visible light, such as the sun or incandescent bulbs, this type of vision system is generally unable to "see" the world. However, at other wavelengths, objects are able to illuminate themselves. According to *Planck's law*, every object above absolute zero emits thermal radiation [39,52]. Most of this radiation falls in the infrared spectrum range (between 0.9 and 14 μm) rather than the visible spectrum range (between 380 and 780 nm). It can be captured by thermal cameras but not by RGB cameras. Therefore, we propose to use thermal imaging as an inexpensive, contactless sensing modality to recognize

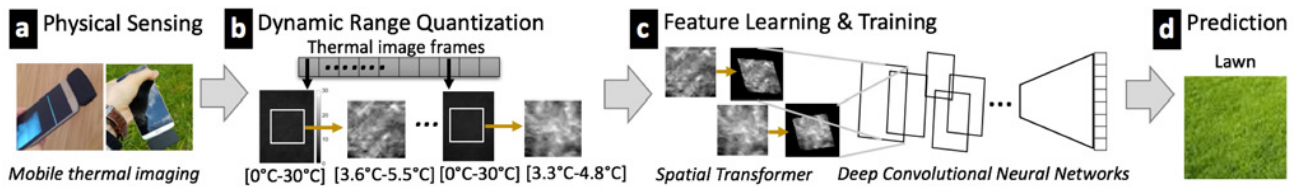

Figure 2. The overall process for proximate material type recognition with Deep Thermal Imaging: a) physical sensing via mobile thermal imaging, b) dynamic range quantization on every thermal image frame, c) feature learning and training process based on spatial transformer and deep convolutional neural networks, d) prediction results.

material types independently of the level of illumination (e.g., working in dark environment).

**Thermal Imaging & its HCI Applications**

Thermal imaging is a key method for measuring spatial temperature of materials, objects and scenes [36]. It has been used in applications that range from inflammation diagnosis [41] to vehicle occupant detection [39], and to breathing and stress monitoring [8,10]. Thermal imaging has also been investigated for HCI applications. Larson et al. [32], for instance, have proposed a thermal imaging-based surface gesture recognition system which uses residual heat traces on a surface. Kurz [30] extended this concept to support touch detection in 3D augmented space. Shirazi et al. [47] have shown how to use thermal reflection characteristics of materials for gestural interaction. However, these approaches have had limited use in the wild because they used high cost, heavy thermal cameras and were sensitive to environmental temperature changes.

Recently, a new group of low-cost, lightweight (<35gram) and mobile thermal cameras have been developed (see Table 1). They have the potential to deploy and power thermal imaging based HCI applications (a smartphone embedding a thermal camera has been already launched [61]). Indeed, Naik et al. [37] have shown that, in controlled environments, a low-cost thermal camera has sufficient resolution and accuracy to grade the maturity of a fruit. Furthermore, Cho et. al [10] have recently investigated how to address the effect of high temperature changes on low cost, low resolution thermal cameras. They proposed an optimal quantization method to compensate for high thermal dynamic range scenes (e.g., outdoor environments with highly varying temperature). The method, together with novel machine learning techniques, may enable the use of low cost thermal camera in more ambitious and unconstrained material recognition applications.

In the next section, we introduce our new mobile thermal imaging-based proximate material type recognizer that can work in the wild in a wide range of environmental illumination, temperature, and view point.

**MATERIAL RECOGNITION STRATEGY**

Deep Thermal Imaging is a novel approach for sensing and recognizing the material of proximate objects. Rather than focusing on the overall material temperature, we propose to use thermal cameras to capture the characteristic spatial thermal pattern of the compounds or elements forming a material [4]. Figure 2 illustrates our four-stage approach: a) physical sensing of proximate surfaces through mobile thermal imaging; b) dynamic range quantization to decrease environmental temperature effects and enhance thermal texture; c) deep-learning-based feature learning and training; and d) real-time recognition of materials.

**Physical Sensing: Mobile Thermal Imaging**

In our study, we used a mobile, low-cost thermal camera (FLIR One 2G, see Table 1) which converts 8-14μm radiation spectral density to temperature. When attached to a smartphone, it can be used to continuously sense and record thermal image sequences of different types of surface materials. Its portability means that it is easy to gather images of the same object from many different perspectives (e.g., shape, geometry, illumination) and distances.

**Dynamic Range Quantization**

Various factors affect the thermal texture of materials captured by the camera. Each material's compound combinations, surface structure (e.g., roughness) and geometry (e.g., cavities) influence both the amount of radiation emitted from the surface and the emissivity [52]. Given this, thermal imaging detects the thermal radiation and *estimates* the temperature by using an emissivity value, so-called *apparent temperature* (not *true temperature)*. Instead of attempting to measure the true temperature of the material, we aim to capture the thermal radiation differences over material surfaces to discriminate them.

| Product | Spatial Resolution | Frame Rate | Thermal Sensitivity (NETD) | Price (in July 2017) |
|---|---|---|---|---|
| [+]FLIR One 2G | 160x120 | <8.7Hz (unsteady) | <0.10°C | Not applicable |
| [+]FLIR One 3G | 80x60 | <8.7Hz (unsteady) | 0.15°C | $199.99 |
| [+]FLIR One Pro | 160x120 | <8.7Hz (unsteady) | 0.15°C | $399.99 |
| [+]Seek Thermal Compact | 206x156 | <9Hz (unsteady) | 0.50°C | $249.00 |
| [+]Seek Thermal Compact PRO | 320x240 | >15Hz (unsteady) | Unknown | $499.00 |
| [+]Therm-App Hz | 384x288 | 25Hz | <0.07°C | $999.00 |
| *FLIR A35 | 320x256 | 60Hz | <0.05°C | $5,900.00 |
| *OPTRIS PI-640 | 640x480 | 32Hz | 0.075°C | $11,310.04 |
| *FLIR A325sc | 320x240 | 60Hz | <0.05°C | $12,322.46 |
| *FLIR SC5000mb | 640x512 | 100Hz | 0.017°C | Over $13,000.00 |

Table 1. Specification and price of commercial thermal cameras: [+]Low-cost, Mobile Thermal Cameras (< 34x14x67mm$^3$ Dimension), *High-end thermal imaging systems (> 90x46x56 mm$^3$ dimension).

To maximize the differences between materials and compensate for environmental temperature changes, we introduce the *Dynamic Range Quantization* method, simplified from the *Optimal Quantization* method proposed in [10]. Quantization, the process of mapping input data to output data within a specific range, is crucial for handling thermal image sequences in the wild [10]. Our method amplifies the *relative* difference between neighboring temperature elements by taking into account the different amount of radiation on every single thermal image of a material surface (Figure 2b).

First, it crops an NxN square at the center of the captured image. In our implementation, we use a 75x75 square given the low spatial resolution (160x120) of the thermal camera. Cropping is important because low-cost systems, designed for mobile thermal imaging, suffer from lens-inducing calibration errors that can lead to extremely low or high temperature at the edge of the image [10,61]. The second step is to adaptively quantize the thermal measurements using the expression $u = \Gamma(T)$, where $[T_1, T_2]$ is the range of temperature and $[u_1, u_2]$ is the range of a pixel value on the output thermal image. The temperature range is given by

$$T_1(k) = \min_{\forall i,j \in ROI} S_{ij}(k), \quad T_2(k) = \max_{\forall i,j \in ROI} S_{ij}(k) \quad (1)$$

where $S_{ij}(k)$ is the element $(i, j)$ of the thermal matrix $S$ with frame number $k$. The final quantized value $I_{ij}(k)$ on the processed thermal image is:

$$I_{ij}(k) = (u_2 - u_1) \cdot \left( \frac{S_{ij}(k) - T_1(k)}{T_2(k) - T_1(k)} \right). \quad (2)$$

This conversion has two effects. First, it produces an output image with more distinct spatial thermal patterns of material (Figure 2b), which help to discriminate between the different types. Second, these patterns are more invariant to changes in ambient temperature.

**Feature Learning Process through Deep CNN**
Given the parallel to image processing with RGB images, our feature learning process uses the deep Convolutional Neural Network (CNN) illustrated in Figure 2c. Many CNNs have been proposed for RGB-vision analysis (e.g., [11,26,40,53]). In this paper, we use the Spatial Transformer-based CNN [26]. This architecture gets its name from the fact that the Spatial Transformer artificially manipulates spatial data information within CNNs. This helps to handle high-level variances of spatial thermal image patterns due to different shooting perspectives and geometries. This technique also performs well on lower resolution images, which is ideal for our thermal image. For more details on the implementation, see [26]. Once the implemented deep learning network has been trained, it can be used to automatically classify material types (Figure 2d). Detailed parameters about the CNN layers are introduced in each study section.

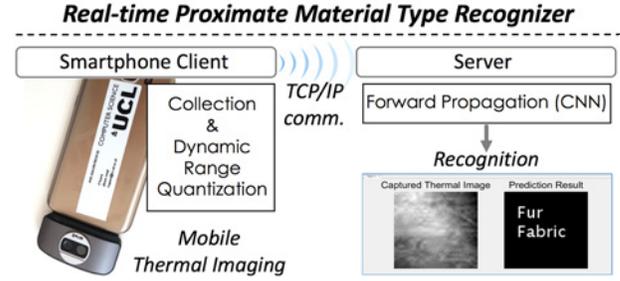

Figure 3. Implementation of Deep Thermal Imaging, the real-time proximate material type recognizer.

**Implementation of Real-Time Proximate Material Type Recognizer**
Figure 3 illustrates the architecture of real-time material recognizer. The smartphone client application (i) controls the mobile thermal imaging sensor (FLIR One and its SDK) to collect a temperature matrix of material, (ii) quantizes the matrix, and then (iii) wirelessly sends it to a server using the TCP/IP protocol. The received data is fed to the trained deep CNN for classification. This transformer network was implemented using the *MatConvNet* framework [51]. Alternatives based on embedded platforms with a GPU could be also used, such as *NVIDIA Jetson TK1* [62] with *Caffe* [63]. A smartphone could also be used for classification by using *TensorFlow Mobile* [64].

Finally, the easy-to-use API for the quantization and the deep learning process and example codes for the raw thermal data collection from the thermal camera are made available at http://youngjuncho.com/2018/DeepThermalImagingAPI.

**MATERIALS IN THE WILD: OUR DATASETS**
There are many possible ways to categorize materials; in our study, we grouped them into *indoor materials* and *outdoor materials*. Figure 4 illustrates the example classes of material surfaces we have targeted. Indoor materials are encountered in the context of a house or an office where the environment is typically more stable and controlled. The selection of indoor materials was mainly based on recent works [48,56], but was extended to include new classes to understand the limits of the approach (e.g., bubble wrap and paper towel given similarity with freezer polyethylene bag and paper, respectively). Outdoor materials are found in conditions which are very variable and are exposed to dirt and weathering effects. The choice of materials was linked to one of our use cases on wheelchair users discussed later. Caretakers were informally interviewed, and prior studies on wheelchair navigation were used (e.g., [27]). Materials which frequently appear on the ground (e.g., granite) or obstacles (e.g., fences) were considered important to facilitate navigation. Such materials are challenging as they are less likely to have constant patterns of spatial temperature, texture and geometry. Two materials were also added to the outdoor collection - *radiator in building corridor (denoted as z)* and *building entrance carpet (denoted §)* even though they are physically found indoors. The reason is that they tend to be more variable – their surfaces may be exposed to high thermal dynamics and dirt.

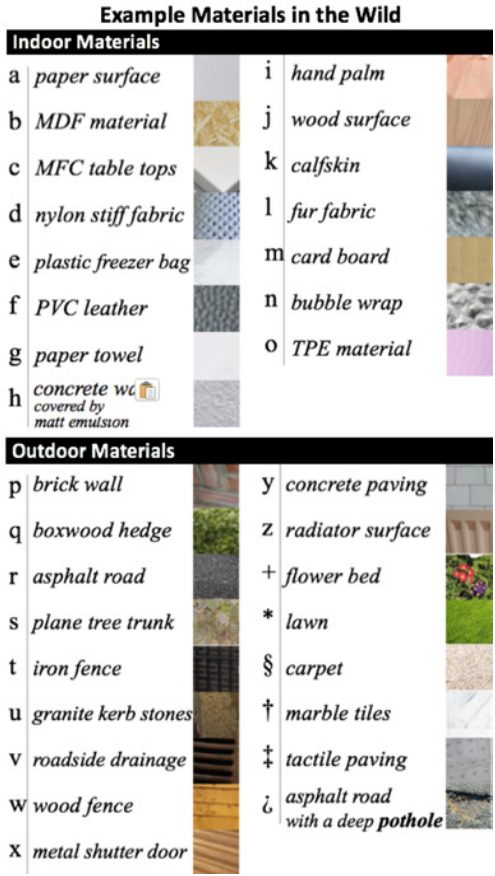

**Figure 4. Example materials collected in the wild: (a) materials generally found in indoor places (less influenced by environmental thermal dynamics), (b) outdoor materials which have richer surface patterns as they are more likely to be affected by surrounding environment temperature, weather and dirt.**

To evaluate the performances of *Deep Thermal Imaging*, we conducted two studies *in the wild* [43] outside of controlled lab settings (e.g., temperature, light and dirt control). The recording of images was carried out by sampling while walking or running around an existing environment and taking thermal videos of material around from different perspective and repeating such recording in various places, at different times of the day, temperature and weather conditions. This process was facilitated by the low and unsteady frame rate of the thermal camera (3-8 thermal images every second), and ensured a collection of non-similar images of the same materials to maximize the generalization capability of the trained recognition model.

### STUDY I: INDOOR MATERIALS

In this section, we report the experimental protocol, our collected thermal image dataset and the recognition results for indoor materials.

### Experimental Protocol & Dataset (DeepTherm I)

For Study I, the 15 indoor materials shown in Figure 4 were used to create the dataset *DeepTherm I* which consists of 14,860 processed thermal images (average count of data for each individual class: 990.7, SD=425.9; 400-600 images of each material per each variable). The dataset was created by recording thermal image sequences in a room with different lighting levels (bright / dark), with/without air-conditioning, different places (on a floor or a desk) and from different perspectives (Figure 5). The spatial temperature patterns were collected from different angles and different distances (between 10 and 50 cm, from the camera lens to the material). The data was collected five times in about 3 weeks. For *i. hand palm* (Figure 4), we collected thermal images of both hand palms from 4 people. Figure 6 shows some of example thermal images collected and quantized for this study. The dataset was divided into a training set and a test set along with a cross-validation method (Set size: 80%-20% in 5-fold; 90%-10% in 10-fold).

### Implemented CNN and Training

The CNN architecture for Study I used the architecture in [26] and consisted of one average pooling layer, two convolutional layers, two max pooling layers and two fully connected layers. We used a Rectified Linear Unit (ReLU) as an activation function to speed up the feature learning process [38]. Quantized images of size 75x75 were resized to 60x60 to be feed-forwarded into the first layer to use the layer composition in [26]. In the training process, a learning rate of 0.001 and a batch size of 256 were used for 350 epochs. The CNN was implemented on the *MatConvNet* framework and ran on a 64-bit Windows desktop (Intel Core i7-3770 CPU @ 3.40GHz 20GB RAM).

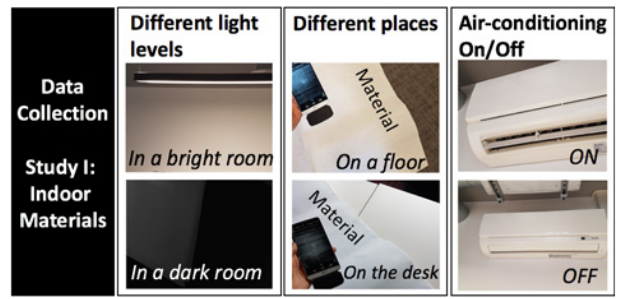

**Figure 5. Data collection for Study I: thermal images of indoor materials were recorded under different conditions (lighting, place, air-conditioning, direction and distance).**

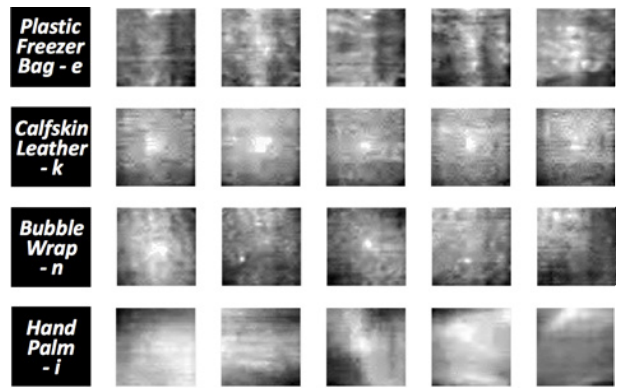

**Figure 6. Examples of collected and quantized thermal images of e. plastic freezer bag, k. calfskin leather, n. bubble wrap and i. hand palm.**

**Results**

Figure 7 shows thermal dynamic levels of the collected thermal images of the 15 indoor materials. The thermal dynamics were examined by using the temperature range $(T_2-T_1)$ in Equation (1). Some materials, such as *i. hand palms* and *n. bubble wrap*, showed higher thermal dynamics than materials that have more regular surface patterns including *a. paper surface* and *b. MDF materials*. The thermal dynamics ranged between 0.259°C and 3.507°C (denoted as *A* in Figure 7) with a mean of 0.889°C (SD=0.4608). We evaluated the performance of the proposed approach on the *DeepTherm I* dataset. We tested using two types of cross-validation (5-fold and 10-fold) within the dataset. The reason was to understand the effects of training with smaller datasets. For the creation of random partitions (i.e., k-fold), the MATLAB function *cvpartition* was used. As results, the CNN classifier achieved a mean accuracy of 97.899% (SD=2.105) and 98.076% (SD=2.075) for the 5-fold and the 10-fold validations respectively. Figure 8 summarizes the mean class accuracy results from 10-fold cross-validation. As it can be seen here, performances are very high for each material type. The worst performing materials were materials *e. plastic freezer bag, n. bubble wrap and o. TPE material*. Even in these cases, the materials were correctly classified in more than 90% of the cases. One explanation possible for these slightly lower performances is that these three types of material are transparent, shiny and reflective and hence their thermal radiation patterns are likely to be more influenced by surroundings [52].

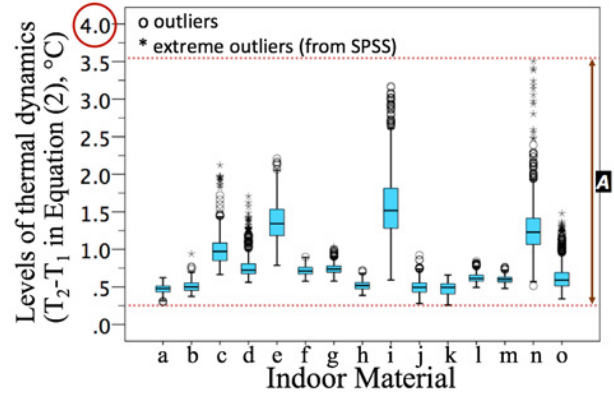

**Figure 7. Levels of thermal dynamics of the total 14860 thermal images across the 15 indoor materials (a-o). The thermal dynamics were estimated by using $(T_2-T_1)$ in Equation (2). [A] indicates the range of those values.**

**STUDY II: OUTDOOR MATERIALS**

This study consists of two parts. The first part follows the design of Study I to investigate the possibility of discriminating between different types of *outdoor* material. The second part investigated the ability of the already trained model to discriminate between the same type of outdoor materials but gathered at real-time and under completely different conditions from the data in the training set (i.e., similar to the *zero-shot learning* [44] testing concept).

*PART I: Training and Testing within the Dataset*

**Experimental Protocol & Dataset (DeepTherm II)**

The 17 outdoor materials (illustrated in Figure 4) were targeted. The data collection process produced the *DeepTherm II* dataset which includes 26,584 labelled thermal images. The average number of collected spatial thermal patterns from each material was 1563.8 (SD=295.3; about 300-500 images of each material per each condition). The accompanying preview video shows examples of data.

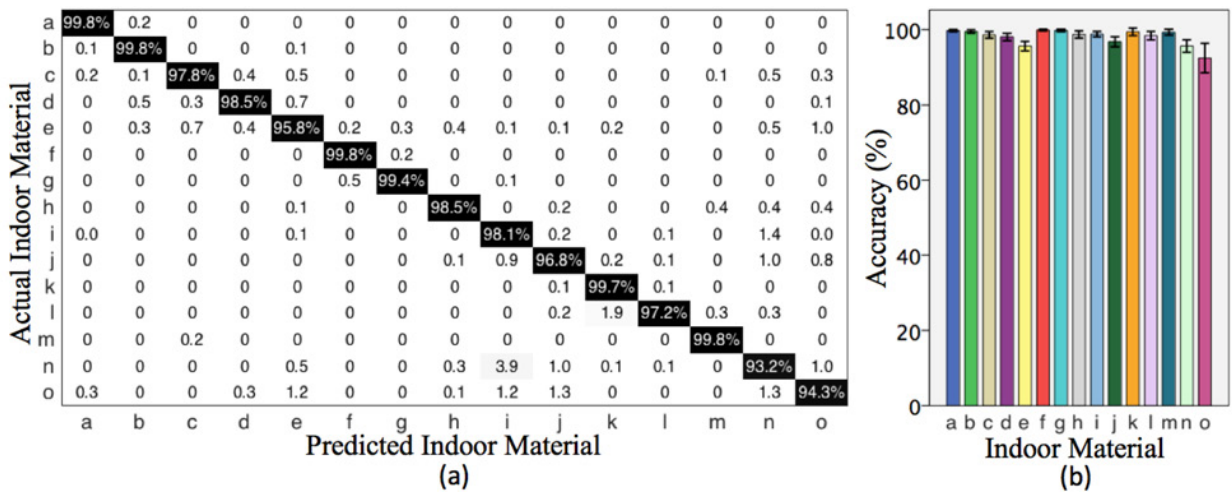

**Figure 8. 10-fold cross validation results from Study I (*low dynamic proximate objects, labels for materials shown in Figure 4*): (a) confusion matrix (in percentage; color range: white (0%) - black (100%)), (b) mean class accuracy distribution (error bar: 95% confidence interval).**

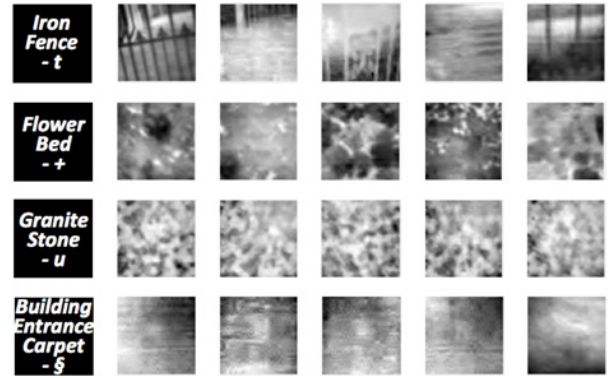

**Figure 9. Data collection for Study II: thermal images of outdoor materials were recorded by two experimenters while walking/running outside both at daytime and nighttime in different seasons.**

As illustrated in Figure 9, two experimenters collected thermal images in two different months (April and August) both during the day and at night. As in the protocol of Study I, experimenters collected spatial material temperature patterns while moving or running slowly around material during shooting to exploit the low frame rate of the thermal camera and recorded materials from different perspectives. As for Study I, imaging materials from different perspectives, distances, times of days, and seasons was critical to evaluate the generalization capability of the modelling approach. However, we avoided rainy days and materials which were significantly covered by clutter such as sand or big trash bags. Figure 10 shows examples of thermal images collected and quantized for this study. A higher number of recordings were collected for materials – such as iron fences – with high degree of variability due to thermal reflection [52].

**Implemented CNN and Training**

A pilot test showed that the CNN architecture used for the first study performed worse on the *DeepTherm II* dataset (less than 70%). A possible reason is that the trained neural network overfitted the training set given noise in the recordings of the outdoor materials. An overfit model is more likely to produce a lower recognition accuracy on new unseen data [50]. Hence, we added "dropout" to the end of the neural network layers in the CNN, which is a simple but powerful way to reduce overfitting issues [29]. The architecture was trained and tested on the DeepTherm II dataset (see parameters in Table 2).

**Results**

Figure 11 shows the thermal dynamic ranges of the collected images of the 17 outdoor materials. The temperatures ranged between 0.341°C and 28.744°C with a mean 2.959°C (SD=1.963). Using 5-fold and 10-fold cross validation methods, the implemented CNN classifier (Table 2) achieved a mean accuracy of 84.526% (SD=9.875) and 89.356% (SD=6.833), respectively. Figure 12 summarizes the mean class accuracy results from 10-fold cross-validation. Surprisingly, *z (radiator in building corridor)* and *§*

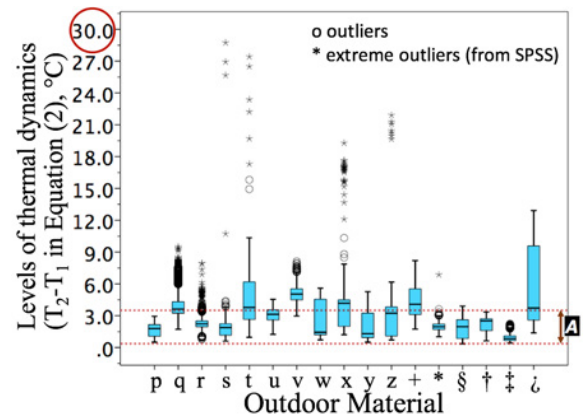

**Figure 10. Examples of collected and quantized thermal images of t. iron fence, +. flower bed, u. granite stones (pavement edge) and §. building entrance carpet. These images contain some degree noise, e.g., dirt and rubbish.**

**Figure 11. Thermal dynamics levels for the total 26584 thermal images across all 17 outdoor materials (p-¿). The thermal dynamics were estimated by using $(T_2-T_1)$ in Equation (2). The thermal dynamic range of indoor materials are marked in red dot lines (*A*) for comparison.**

*(building entrance carpet),* which are on the border between outdoor and indoor areas, showed highest accuracies (97.9% and 98.1%, respectively). On the other hand, *+ (flower bed), * (lawn)* and *¿ (asphalt road with a deep pothole)* showed worst performances just below or around 80%. Those results can be explained by the fact that there is a greater variability in the composition of these materials (e.g., soil, flowers).

*PART II: Testing under Different Conditions in Real-Time*

**Experimental Protocol**

Finally, we assessed the scalability of the Deep Thermal Imaging and its generalization to new unseen data with different conditions, i.e., training on the outdoor dataset of study II and testing on a new dataset containing the same type of material but collected under different conditions, e.g., iron fences *in front of windows* (transparent materials like glass could influence temperature patterns [47]), *rubber covered* tactile paving. In addition, we tested how well the system could recognize material that was partially altered and covered by other material, i.e., pavement *covered by sand*, and *wet because of rain*. This part of the study was conducted in real-time to simulate its use in real-life applications: shooting a thermal image of material and immediately testing the classifier pre-trained from *the DeepTherm II* dataset. The real-time proximate material type recognizer (shown in Figures 1c and 3) was designed to make a prediction when a decision probability was higher than a set threshold value (i.e., a score produced from the softmax function in Table 2).

| Layer | 0 | 1 | 2 | 3 | 4 | 5 | 6 | 7 | 8 | 9 | 10 | 11 | 12 |
|---|---|---|---|---|---|---|---|---|---|---|---|---|---|
| Type | Input | Aver. Pool. | Conv. | ReLU | Max Pool. | Conv. | ReLU | Max Pool. | Fully conn. | ReLU | Fully conn. | **Drop out** | Soft max |
| Kernel size | n/a | 2x2 | 7x7 | n/a | 2x2 | 5x5 | n/a | 2x2 | 1x1 | n/a | 1x1 | (0.3) | n/a |
| Kernel Num. | n/a | n/a | 12 | n/a | n/a | 24 | n/a | n/a | 48 | n/a | 17 | n/a | n/a |

**Table 2. CNN architecture used for Study II: based on [26] and dropout technique [29] to handle overfitting issues.**

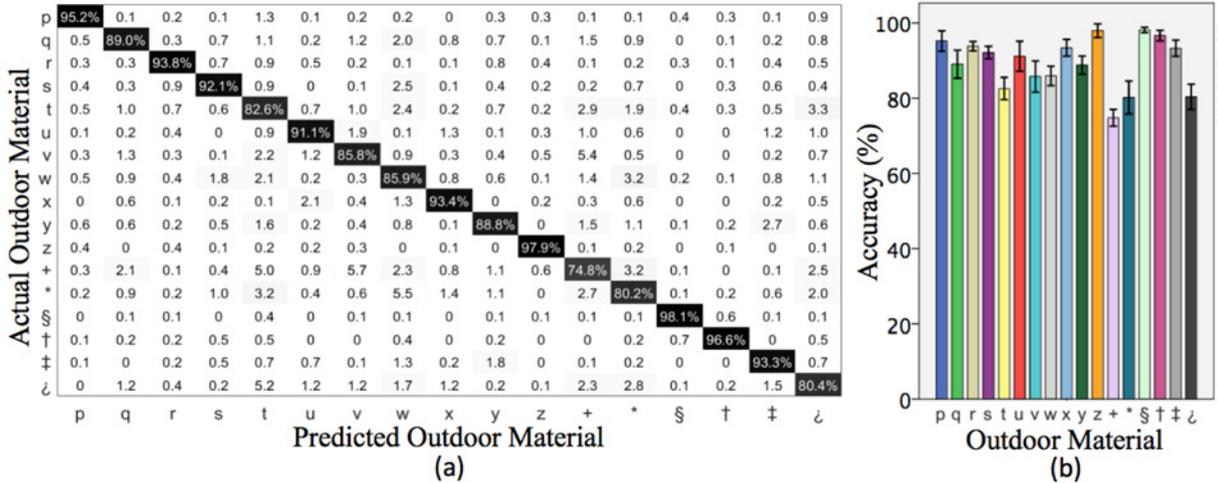

**Figure 12. 10-fold cross validation results from Study II (*high dynamic temperature range proximate material, labels for materials shown in Figure 4*): (a) confusion matrix (in percentage; color range: white(0%)-black(100%)), (b) mean class accuracy recognition distribution (error bar: 95% confidence interval).**

### Results
Figure 13 shows the mean class accuracy results for the thermal images of the new data (in unseen conditions) with respect to the 17 classes from the training set (in Part I of Study II). Prediction results were well above chance level (classifying material among 17 material types) except for the wet cases in comparison with the related dry conditions (e.g., dried asphalt road, concrete paving with clear borders). Performances in recognizing the new iron fence and the rubber-covered tactile paving among 17 outdoor materials were relatively high with 63.16% and 69.23% accuracies, respectively. Given that even the state-of-the-art deep learning techniques for RGB-vision-based material recognition tasks produced about a 50% mean accuracy in discriminating materials across 10 categories from cross-validated dataset experiments (i.e., when training on a dataset and testing on new dataset) [2], the obtained performances can still be considered very good. On the other hand, the classifier trained on *DeepTherm I*I was unable to recognize wet materials, particularly tactile and concrete paving in the 6th and 7th columns in Figure 13, suggesting the need to include such weather condition in the training set.

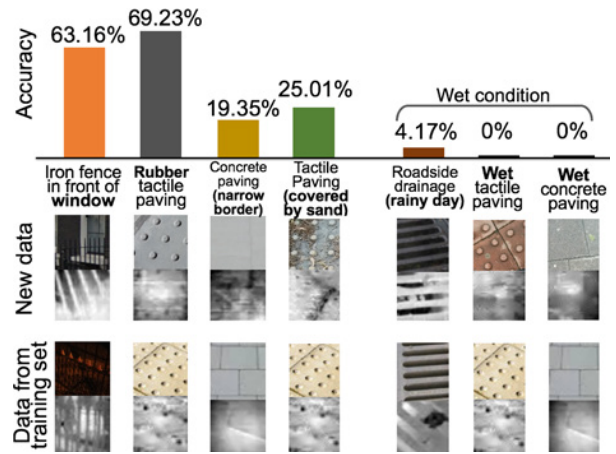

**Figure 13. Mean class accuracy results (17 classes) from real-time recognition test on new data in different conditions (Study II, Part II): prediction results were by and large lower, but we note that this task is highly challenging in image classification (e.g. about 50% mean accuracy from 10 classes cross-dataset experiments in [2]).**

## EXAMPLE USE CASES
Before discussing the lesson learnt from the experimental studies, we discuss some of the possible HCI uses of our *Deep Thermal Imaging* system and its value in comparison with other material-sensing approaches.

### Extension of Indoor Use Cases
As addressed in [48], a material recognizer helps lessen the need for special markers and tags in Augmented Reality

(AR), context-awareness applications. Earlier recognizers, however, limit possible scenarios due to the need to be in contact with the material to be recognized or to use special surfaces (e.g., adding transparent material to a showcase as a tag to provide video instruction on a user's smartphone [48]). *Deep Thermal Imaging* does not have such requirements allowing more seamless HCI applications. For example, by integrating a thermal camera into a AR headset (e.g., Microsoft Hololens), a home redecoration application could help its user select the right type of painting for the walls or for any other object in the home by simply moving around the home. The application could show how that surfaces would appear when painted with a particular color of a certain type of paint. Indeed, the resulting color depends on the material properties and not just on the paint used. A simpler version could be obtained by using a smartphone rather than the AR headset by pointing the smartphone to the objects as the person moves around.

Another interesting example could be the improvement of human assistant robots, such as, vacuum cleaning robots (e.g., Dyson 360 Eye robot [65], LG Hom-Bot square [66]). These systems use ultrasonic sensors [33] and photo-detectors [21] to detect the presence of obstacles. However, these systems are limited in their recognition of floor types. Users must intervene to adjust parameters and tooling for maximizing the task performance when changing floor type (e.g., hard marble floor to deep pile carpet as shown in Figure 14(d)). Given that our system is less affected by ambient illumination and can differentiate between material types in a contact-less manner, it can facilitate automatic tools and parameters adjustment without the continuous intervention of a human operator. The capability could be extended to other floor types beyond the tested floor materials (e.g., marble, carpet) by increasing the training set.

**Outdoor Use Cases**

Deep Thermal Imaging could become an artificial eye to increase a person's ability to discern outdoor surroundings without touching them. Visually impaired people could benefit from this approach. As discussed earlier, the Microsoft *Seeing AI* [58] can be of a huge help for the community with its function to describe the scene around a user. Beyond typical situations with moderate light levels, a smartphone RGB camera used for the Seeing AI may not work properly. The performance could be improved by complementing it with our system that is illumination-independent.

A wheelchair user could also exploit this system to avoid more unpleasant footpaths when navigating at night or backwards and in crowded areas with the setup described in Figure 14(a). Informal interviews with wheelchair users highlight the difficulty that footpaths may present. Deep Thermal Imaging integrated on the wheelchair could detect poor surfaces as the wheelchair approaches it (e.g., uneven surface, manhole cover [27]) and alert the user (e.g., through sound, vibration) or inform the automatic engine driver. It can also help caretakers when pushing a wheelchair given their limited visibility of the footpath (as per interviews).

Another important application, extending the previous one, is road material and state crowdsourcing [19,24,25]. This is fundamental to improve route planning for pedestrians, cyclists including the user groups described above. This is particularly important in a novel environment or in non-industrialized countries where streets are not yet well mapped or maintained [19,23–25]. Crowdsourcing, through the setup (with a smartphone GPS) in Figure 14(a), could help build a smart city map (e.g., Figure 14(b)) by improving the information provided by current maps (e.g., Google Accessibility Map [67]). Similarly, the crowdsourcing application could help detect potholes on roads which cause serious problems in metropolitan cities like London and New York [17,46]. Lots of money and time are required to find and repair potholes. Beyond voluntary tagging, automatic sensing using accelerometer and GPS is being currently explored for this purpose [17]. However, the information extracted from such approach contains vibration noise. Our system could be used to complement the information provided by the accelerometers to further confirm the presence of a pothole and also indicate the pavement material for preliminary road maintenance planning (Figure 14(c)).

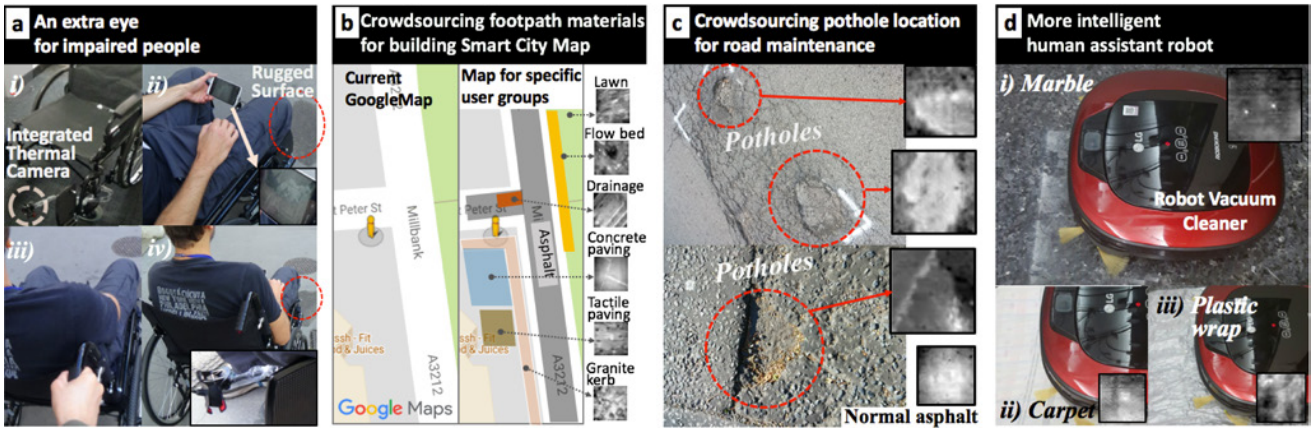

Figure 14. Example applications: (a) an additional eye for people (visually impaired people or wheelchair users could benefit by being alerted that a footpath is wet or in poor conditions even at night), (b) crowdsourcing materials on footpath for building smart city maps, this in turn can help bicycle users, runners, the low vision community to navigate using route most suitable to their means and needs, (c) crowdsourcing pothole locations for road maintenance (potholes cause a serious problem in cities, maintaining them is challenging), and (d) more intelligent human assistant robot (Deep Thermal Imaging can provide the ability to recognize the floor types and conditions, leading to more efficient and adaptive smart-home robots).

## DISCUSSION

Novel technological sensing approaches are central to many HCI innovations. Recent work [2,48,53,55] investigated material recognition to support material awareness. Building on them, our paper contributes to this body of work by proposing mobile thermal imaging as a new recognition modality to supplement some of the limitations of existing approaches. Our approach, in particular, offers higher flexibility when used in the wild where there are a variety of materials and environmental factors. It does not require contact between the sensor and the material, reducing safety concerns, and social issues and creating more seamless types of interactions and applications. It is accessible to general CHI audiences, hobbyists and end users (e.g., impaired people) as it works with low-cost hardware and uses a deep learning-based approach which can be easily trained to recognize new materials without requiring any special skills by the end user (i.e., feature hand-crafting).

Two main studies were conducted to investigate the efficacy and potential of the Deep Thermal Imaging system in recognizing typical indoor and outdoor material under different illumination, temperature conditions and point of views. Overall, Deep Thermal Imaging showed highly good performance in discriminating between 15 indoor materials and between 17 outdoor surfaces types. The mean class accuracies were slightly higher in the case of indoor materials than outdoor ones. The slightly lower performance for the outdoor material can be explained by higher thermal dynamics (see Figures 7 and 11) and lower homogeneity between objects of similar material. For example, different parts of a pedestrian path could have different amounts of dirt or footprints, altering surface thermal radiation patterns.

### Limitations and Future Directions

Despite the high performances shown by our results, there is space for improvement. First, thermal imaging is affected by thermal reflection of wet and glossy materials [28,47,52]. For instance, the reflective bubble wrap and plastic freezer bag in Figure 6 (material *n* and *e*) had some sporadic (high apparent temperature) white spots on their thermal images. This is due to their high thermal reflectivity [28,47,52] leading to the higher thermal dynamic ranges in Figure 7. In the case of completely unseen contexts and their interaction (e.g., Study II/Part II), the performances also decreased (as expected) but they are still well above chance level. This suggests that various environmental factors such as weather and adjacent materials could also influence its ability to generalize. Given that reflectance and transmission are influenced by thermal radiations received from other materials, material with high reflectivity and transmissivity may require a larger dataset as they may be more affected by the environments and material they are placed on or closed to. Given the nature of deep learning, growing a dataset and optimizing parameters of the model [26] could help generalize to the variety [18,57]. A possible way to continuously increase the dataset is to build a crowdsourcing site as discussed for the various HCI use cases.

Lastly, constraints arise from the resolution and imaging quality of the thermal imaging system itself. In particular, low spatial resolution limits the distance at which materials can be classified to less than 0.5 meter. In addition, the thermal sensitivity (<0.1°C) was also limited. Some quantized thermal images in Figure 6 (material *k*) and Figure 10 (material *ş*) have a white round spot at the center which were possibly induced by the limited sensitivity. When we relate it to their narrow range of temperature distribution (lower than 1°C, see Figure 7 and 11), the bulk of thermal radiation differences within the spot might not be captured. When we first undertook this Deep Thermal Imaging project, only the *FLIR One (2G)* was available as a mobile thermographic channel. However, now we can see more options coming to the market such as *Therm-App Hz* (see Table 1). Given the high performances produced by our method, we can expect that by using the new low cost but more sensitive cameras that are emerging, these performances will get even better. Indeed, our approach is applicable to any types of thermal cameras. Beyond this unimodal sensing, combination of multiple low-cost mobile sensors could be the next step to build better material and context awareness (e.g., LIDAR rangefinder to easily detect the depth of potholes and our system to do material types).

## CONCLUSION

Deep Thermal Imaging is a new technique to recognize materials in proximity which is based on the use of mobile thermal imaging and deep learning. To the best of our knowledge, this paper provides the first demonstration that spatial thermal patterns of materials can be used to enable automatic material recognition tasks with good performances (above 89% mean accuracy from 15 indoor and 17 outdoor materials). We leveraged a mobile, low-frame rate thermal camera to build a large dataset (41,444 images) of variant thermal textures of materials. Potential use cases of the systems compared with the state of the art are also discussed highlighting the potential of the system. The collected DeepTherm I and II datasets and API are available for download at (http://youngjuncho.com/datasets).

## ACKNOWLEDGMENTS

Youngjun Cho was supported by University College London Overseas Research Scholarship (UCL-ORS) awarded to top quality international postgraduate students.